\documentclass[lettersize,journal]{IEEEtran}
\usepackage{amsmath,amsfonts}
\usepackage{algorithmic}
\usepackage{algorithm}
\usepackage{array}
\usepackage[caption=false,font=normalsize,labelfont=sf,textfont=sf]{subfig}
\usepackage{textcomp}
\usepackage{stfloats}
\usepackage{url}
\usepackage{verbatim}
\usepackage{graphicx}
\usepackage{cite}
\usepackage{amssymb}
\usepackage{booktabs}
\usepackage{multirow}
\usepackage{upgreek}
\usepackage{etoolbox}
\usepackage{caption}

\begin{document}

\title{MMFace4D: A Large-Scale Multi-Modal 4D Face Dataset for Audio-Driven 3D Face Animation
}

\author{Haozhe Wu, Jia Jia, Junliang Xing~\IEEEmembership{Senior Member,~IEEE,}, Hongwei Xu, Xiangyuan Wang, Jelo Wang}

\markboth{IEEE TRANSACTIONS ON MULTIMEDIA}%
{Shell \MakeLowercase{\textit{et al.}}: A Sample Article Using IEEEtran.cls for IEEE Journals}


\maketitle



\begin{abstract}
Audio-Driven Face Animation is an eagerly anticipated technique for applications such as VR/AR, games, and movie making. %
With the rapid development of 3D engines, there is an increasing demand for driving 3D faces with audio. %
However, currently available 3D face animation datasets are either scale-limited or quality-unsatisfied, which hampers further developments of audio-driven 3D face animation. %
To address this challenge, we propose \textbf{MMFace4D}, a large-scale multi-modal 4D (3D sequence) face dataset consisting of 431 identities, 35,904 sequences, and 3.9 million frames. %
MMFace4D exhibits two compelling characteristics: 1) a remarkably diverse set of subjects and corpus, encompassing actors spanning ages 15 to 68, and recorded sentences with durations ranging from 0.7 to 11.4 seconds. 2) It features synchronized audio and 3D mesh sequences with high-resolution face details. To capture the subtle nuances of 3D facial expressions, we leverage three synchronized RGB-D cameras during the recording process. %
Upon MMFace4D, we construct a non-autoregressive framework for audio-driven 3D face animation. %
Our framework considers the regional and composite natures of facial animations, and surpasses contemporary state-of-the-art approaches both qualitatively and quantitatively. %
The code, model, and dataset will be publicly available. %
\end{abstract}

\begin{IEEEkeywords}
Dataset, 4D, Multi-Modal, Face Animation.
\end{IEEEkeywords}

\section{Introduction}
\label{sec:intro}

\begin{figure*}[h]
  \centering
  \includegraphics[width=\linewidth]{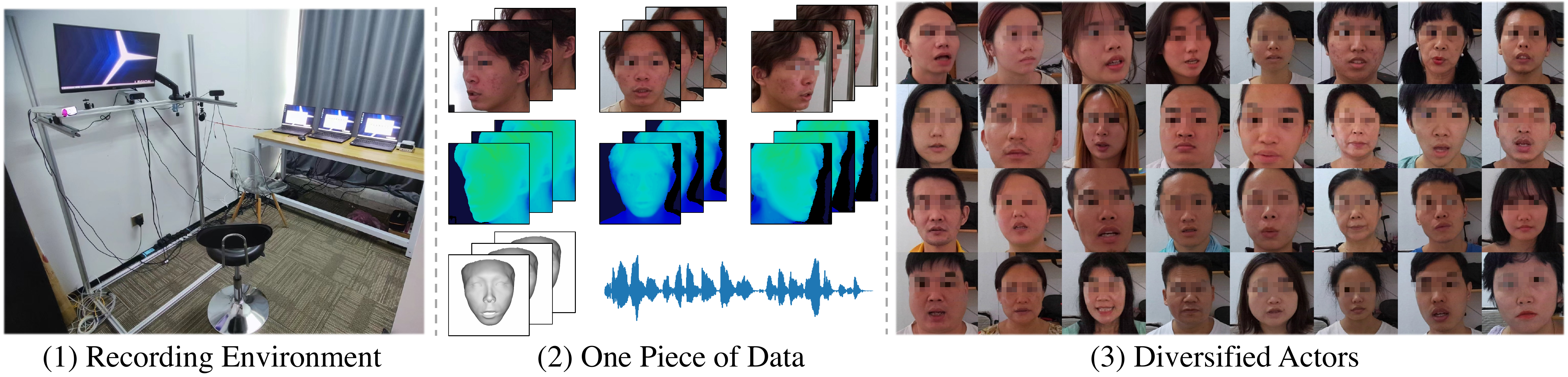}
  \caption{An overview of the MMFace4D dataset. %
    We demonstrate (1) the capture setup and the recording environment, (2) one piece of the MMFace4D data, which contains multi-view RGB videos, multi-view depth videos, reconstructed meshes, and synchronized speech audio, (3) the diversified actors in the MMFace4D dataset. Note that all RGB facial images are mosaicked for privacy protection. 
  }
  \label{fig:intro}
\end{figure*}

The Audio-Driven Face Animation is a fundamental problem in the area of digital avatar synthesis, %
which is essential for several applications, such as VR/AR, games, and movie-making. %
Recently, significant progress has been made with the advent of deep learning and the release of multi-modal face animation datasets. %
The majority of prior works aimed to produce 2D talking head videos due to the availability of massive audio-visual datasets~\cite{afouras2018deep, Nagrani17, Afouras18d, Chung18b, shillingford2018large, yang2019lrw, rossler2019faceforensics++, wang2020mead, zhang2021flow}. %
Whereas, researches in 3D face animation~\cite{richard2021meshtalk, cudeiro2019capture,fan2022faceformer,xing2023codetalker}, which are more closely related to 3D applications like games and movie making, are still in infancy because of the size or quality constraints of currently available datasets. %
Some 3D face animation datasets are constructed with powerful devices~\cite{cudeiro2019capture} and thus have high-fidelity. %
However, the sizes of these datasets are limited. %
Meanwhile, some 3D face animation datasets are constructed on a large scale with monocular 3D reconstruction method~\cite{deng2019accurate}. %
Nevertheless, these datasets have limited quality because the 3D faces are represented as low-dimensional 3DMM coefficients~\cite{blanz1999morphable}. %
The constraints of these datasets impose a hurdle for training generalizable and realistic 3D face animation models. %

To address these issues, we present \textbf{MMFace4D}, a large-scale multi-modal 4D~(3D sequence) face dataset, which captures high-fidelity 3D face sequences and synchronized audios for each actor. %
Figure~\ref{fig:intro} demonstrates the recording environment and one piece of the MMFace4D dataset. %
The MMFace4D dataset comprises 431 identities, 35,904 sequences, and 3.9 million frames. %
More importantly,  MMFace4D features two main properties listed below:

\textbf{- Diversified Subjects and Corpus}. For subjects, the MMFace4D dataset covers diversified actors with median age of 28, minimum age of 15, and maximum age of 68. %
For corpus, we build a corpus with 11,000 sentences under different scenarios such as news broadcasting, conversation, and storytelling. %
The corpus contains 2000 neutral sentences and 9000 emotional sentences of six basic emotional types~(angry, disgust, happy, fear, sad, surprise). %
The recorded sentences have duration ranging from 0.7 to 11.4 seconds. %

\textbf{- High-fidelity Animation and Synchronized Audio}. 
We capture the RGB-D videos of facial movements with three synchronized RGB-D cameras from different views. %
Based on the captured RGB-D videos, we develop a reconstruction pipeline to fuse the multi-view RGB-D videos and obtain topology-uniformed 3D mesh sequences. %
Compared with the 3DMM-based reconstruction methods~\cite{yang2020facescape,blanz1999morphable}, our pipeline preserves subtle motions of the mouth, cheek, and eyebrow. This is achieved through a thoughtful design involving multi-stage fitting and vertex-level optimization. %
Besides, for each 3D mesh sequence, we capture synchronized speech audio with a directional microphone. 

We conduct extensive observations on the MMFace4D dataset, verifying that our dataset has various talking styles, expressive facial motions, and diversified actors. %
These characteristics empower the training of high-fidelity, expressive, and generalizable face animation models. %

Building upon MMFace4D, we introduce a novel audio-driven 3D face animation method, extending our previous work in~\cite{wu2023speech}. %
Our approach addresses two essential facets of facial animations: the composite nature, capturing how speech-independent factors globally modulate speech-driven facial movements temporally, and the regional nature, recognizing that facial movements are locally influenced by musculature spatially. To tackle the composite nature, we introduce an adaptive modulation module for combining speech-independent and speech-dependent movements. For the regional nature, we propose a sparsity regularizer to ensure that each facial feature component focuses on the local spatial movements of 3D faces in every frame. Additionally, we present a non-autoregressive backbone for audio-to-3D facial movement translation, maintaining high-frequency nuances and enabling efficient inference. Extensive experiments on the MMFace4D dataset validate the effectiveness of our framework, surpassing state-of-the-art approaches significantly. %
The code, model, and dataset will be publicly available. %



\section{Related Work}

\textbf{Audio-Driven Face Animation.} Audio-Driven face animation has received significant attention in previous
literature. %
Related work in this area can be grouped into two categories: the 2D-based approaches~\cite{chen2018lip,das2020speech,fan2015photo,ji2021audio,prajwal2020lip,vougioukas2020realistic,zhou2019talking,sinha2022emotion,ye2022audio,wang2022anyonenet,eskimez2021speech,yu2021multimodal} and the 3D-based approaches~\cite{edwards2016jali,taylor2012dynamic,fan2022faceformer,richard2021meshtalk,cudeiro2019capture,liu2021geometry,wu2021imitating,guo2021ad}. %

With the availability of large-scale audio-visual datasets, 2D-based approaches have been paid much attention. %
These methods usually leverage optical flow~\cite{sinha2022emotion}, landmarks~\cite{das2020speech,ji2021audio}, or disentangled representations~\cite{chen2018lip,zhou2019talking} to synthesize realistic audio-driven taking faces. %
The 2D videos generated by these methods have wide applications in movie dubbing, 2D games, and telecommunications. %
However, for 3D applications such as VR/AR, filmmaking, and 3D games, these approaches are less applicable. %

\begin{table}[]
\setlength\tabcolsep{2pt}
\centering
\caption{Comparison of 4D (3D sequence) face datasets, of which each sequence has synchronized speech audio. %
Note that only the publicly available datasets are compared as follows~\cite{fanelli20103,cudeiro2019capture,richard2021meshtalk, wang2020mead}. %
    MMFace4D dataset has a competitive scale in terms of subject number (\#Subj), corpus scale (\#Corp), sequence number (\#Seq), and duration (\#Dura). Additionally, the frame per second (FPS), the emotion label (Emo), the spoken language (Lang), and the presence of topology-uniformed meshes (Mesh) are also listed. }
\begin{tabular}{cc|cccc|c}
\hline
\multicolumn{2}{c|}{Dataset}                            & BIWI    & VOCA    & MeshTalk & MEAD & \textbf{MMFace4D}    \\ \hline
\multicolumn{1}{c|}{\multirow{4}{*}{Scale}}    & \#Subj & 14      & 12      & 250  &  60  & 431     \\ 
\multicolumn{1}{c|}{}                          & \#Corp & 40      & 40      & 50   &  159    & 11,000   \\
\multicolumn{1}{c|}{}                          & \#Seq  & 1109    & 480     & 12,500  & 40000  & 35,904   \\
\multicolumn{1}{c|}{}                          & \#Dura & 1.44h   & 0.5h    & 13h   & 40h   & 36h     \\ \hline
\multicolumn{1}{c|}{\multirow{4}{*}{Property}} & \#FPS  & 25      & 60      & 30   & 30    & 30      \\ 
\multicolumn{1}{c|}{}                          & Emo    & \checkmark       & -   & \checkmark    & -        & \checkmark       \\ 
\multicolumn{1}{c|}{}                          & Lang   & English & English & English & English  & Chinese \\
\multicolumn{1}{c|}{}                          & Mesh   & \checkmark & \checkmark & \checkmark & -  & \checkmark \\ \hline
\end{tabular}
\label{tab:data_compare}
\end{table}

To animate 3D faces, some researchers~\cite {edwards2016jali,taylor2012dynamic} proposed rule-based methods to build a mapping between input audio and rigged 3D faces. %
For example, the JALI model~\cite{edwards2016jali} respectively builds the jaw model and lip model for face animation, the dynamic viseme model~\cite{taylor2012dynamic} captures visual coarticulation and the inherent asynchrony between visual and acoustic speech. %
These rule-based methods usually require intensive manual labor to achieve realistic animation. %
To alleviate the requirement of manual labor, several data-driven methods have been proposed~\cite{fan2022faceformer,richard2021meshtalk,cudeiro2019capture,liu2021geometry,wu2021imitating,guo2021ad,lahiri2021lipsync3d,yao2022dfa,karras2017audio,liu2015video,taylor2017deep}. %
Some methods use monocular 3D face reconstruction to synthesize 3D faces from video data~\cite{wu2021imitating,lahiri2021lipsync3d,liu2015video}. These methods generalize well across different subjects, %
whereas the animation quality of these methods is limited due to the constraints of monocular 3D reconstruction. %
The AD-NeRF and DFA-NeRF~\cite{guo2021ad,yao2022dfa} leverage neural radiance field~\cite{mildenhall2021nerf} to synthesize 3D face animations with granular control on pose and emotion. %
Nevertheless, the NeRF-based methods cannot generate topology-uniformed 3D meshes, which is inapplicable for several 3D applications. %
To synthesize highly-realistic and topology-uniformed 3D mesh sequences, %
researches have explored both speaker-independent ~\cite{xing2023codetalker,fan2022faceformer,richard2021meshtalk,cudeiro2019capture,liu2021geometry} and speaker-dependent~\cite{karras2017audio} frameworks. %
These methods require high-resolution 3D face animation datasets with diversified subjects and corpus. %
However, currently available datasets are insufficient in terms of either diversity or 3D resolution, %
which severely obstructs the training of generalizable and realistic 3D face animation models. %
Such urgent demand brings out our MMFace4D dataset. %


\textbf{3D and 4D Face Datasets.}
Several 3D~\cite{cao2013facewarehouse,yang2020facescape,paysan20093d} and 4D face~\cite{zhang2014bp4d,alashkar20143d,cosker2011facs,zhang2013high,zhang2016multimodal,richard2021meshtalk,cudeiro2019capture,fanelli20103} datasets have been released to analyze static and dynamic facial expressions. %
These datasets record high-resolution 3D faces, %
which are widely applied in 3D face recognition, 3D face morphable model, 3D expression analysis and \textit{et al}. %

Meanwhile, only a few datasets capture synchronized audio with dynamic 3D faces. %
The BIWI dataset~\cite{fanelli20103}, VOCASET~\cite{cudeiro2019capture}, MeshTalk dataset~\cite{richard2021meshtalk}, and the MEAD dataset~\cite{wang2020mead} are mostly used to train audio-driven 3D face animation models. %
The BIWI dataset records 40 spoken English sentences for each of 14 subjects. %
The VOCASET records 29 minutes of 4D scans from 12 speakers. %
Both BIWI and VOCASET are small-scale, which limits the generalization capacity of 3D face animation models. %
The MeshTalk dataset collects 250 subjects, each of which reads 50 phonetically balanced sentences. %
However, the MeshTalk corpus has a limited size, and only a few parts of the MeshTalk dataset are publicly available. %
The MEAD dataset comprises 40,000 sequences captured from 60 actors using 7 RGB cameras, offering various viewing angles. However, it lacks topology-uniformed meshes, presenting a challenge for generating animations at each 3D vertex. Researchers working with the MEAD dataset often rely on NeRF-based methods~\cite{mildenhall2021nerf} for synthesizing 3D facial animations. To overcome these limitations, we introduce MMFace4D, a dataset featuring diverse subjects, extensive corpus data, and topology-uniformed meshes. Table~\ref{tab:data_compare} gives the comparison of all the datasets above. %



\section{Capture Setup}

We devise a capture system comprising three RGB-D cameras, one microphone, and one screen. %
Each camera is placed at the height of 1.2 meters. %
One camera shoots at the front of the face, %
the other two cameras shoot at the left and right sides with 45 degrees of angle. %
We leverage Azure Kinect Camera\footnote{\url{azure.microsoft.com/en-us/products/kinect-dk/}} to capture RGB-D video. We record RGB video with a resolution of $1920 \times 1080$, and record depth video with a resolution of $640 \times 576$. %
The three cameras are synchronized with the daisy-chain configuration, where the synchronization signal is sequentially passed from the master camera to the other two subordinate cameras. 
To accurately align the RGB-D images of the three cameras,  we leverage the point-to-plane ICP~\cite{rusinkiewicz2001efficient}. %
The camera extrinsics are initialized by detected facial landmarks. %
Meanwhile, we leverage COMICA VM20 microphone\footnote{\url{www.comica-audio.com/product/CVM-VM20-82.html}} to record audio. %
Figure~\ref{fig:intro} shows the setup. %

\begin{figure*}[h]
  \centering
  \includegraphics[width=\linewidth]{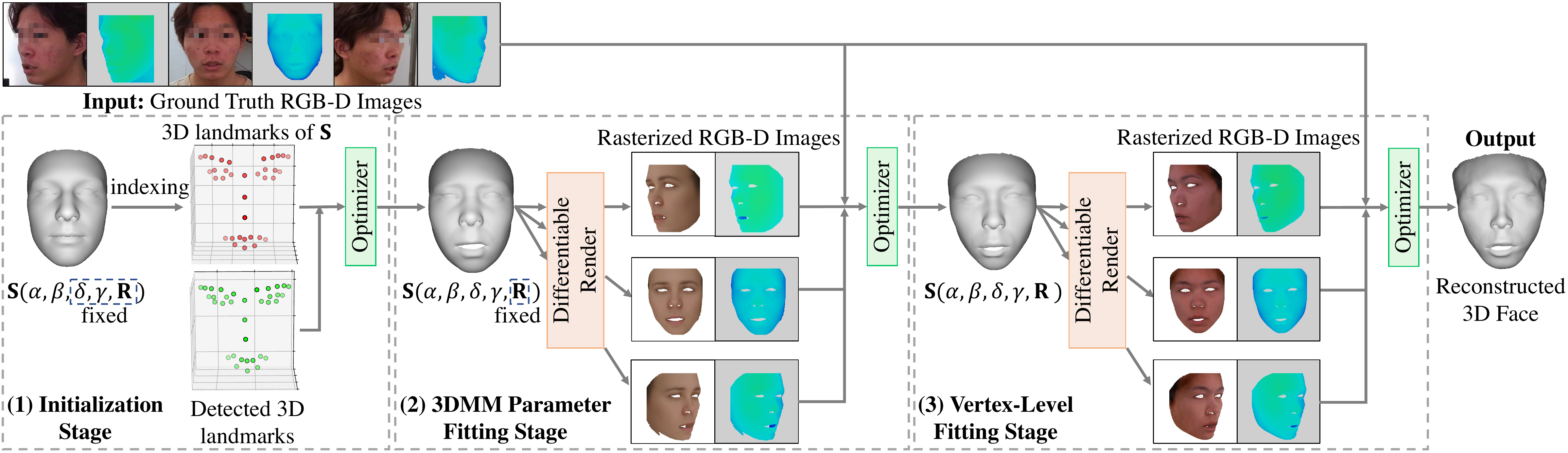}
  \caption{The 3D face reconstruction pipeline with three stages: initialization, 3DMM parameter fitting, and vertex-level fitting. %
  $\mathbf{S}$ denotes the 3D face, $\alpha, \beta, \delta, \gamma, \mathbf{R}$ are respectively shape, expression, texture, lighting, and vertex-level deformation parameters.
  }
  \label{fig:framework}
\end{figure*}


Before recording, we built a large-scale corpus with 11,000 sentences under different scenarios such as news broadcasting, conversation, and storytelling. %
Each sentence has an emotion label of seven categories (neutral, angry, disgust, happy, fear, sad, surprise). %
For the neutral emotion, we have 2000 sentences. %
For the other emotions, we have 1500 sentences. %
Each sentence of the corpus has 17 words on average. %
Our corpus covers each phoneme as evenly as possible. %

Each actor participates in the recording voluntarily. %
The actor is asked to practice 100 sentences of different emotions fluently and emotively. %
Afterward, the capture device starts to record each sentence one by one. %
Note that despite each actor trying to read each sentence with emotion as much as possible, a few sentences are not as expressive as expected because some actors are not professional. %
Nevertheless, most recorded sentences are still expressive. %

\section{Toolchain}

From the capture devices, we have obtained the speech audio and synchronized RGB-D videos from three different views. %
In this section, we will reconstruct topology-uniformed 3D Face sequences from the RGB-D videos. %
We take the basel face model~(BFM)~\cite{paysan20093d} as a template 3D face and deform the template to fit RGB-D videos. %
The overall pipeline is illustrated in Figure~\ref{fig:framework}. %
The pipeline has three stages: (1) initialization stage, (2) 3DMM parameter fitting stage, and (3) vertex-level fitting stage. %

\subsection{Initialization}

Before reconstruction, we first need to define the deformation space of each 3D face and conduct initialization. %

In our pipeline, each 3D face is characterized by two levels of deformation spaces: the 3DMM level, which describes coarse-grained facial shapes and movements, and the vertex level, which captures fine-grained details. %
Formally, given the 3D face $\mathbf{S} \in \mathbb{R}^{n \times 3}$ (where $n$ is the number of vertices), the 3DMM level deformation is represented as an affine model of facial expression parameter $\mathbf{\alpha}$ and facial identity parameter $\mathbf{\beta}$, the vertex level deformation is represented as offset matrix $\mathbf{R} \in \mathbb{R}^{n \times 3}$. The two types of deformation are added together to formulate the full deformation space of $\mathbf{S}$:
\begin{equation}
  \mathbf{S} = \mathbf{\bar{S}} + \mathbf{B}_{id}\mathbf{\alpha} + \mathbf{B}_{exp}\mathbf{\beta} + \mathbf{R},
\end{equation}
where $\mathbf{\bar{S}}$ is the average face shape; $\mathbf{B}_{id}$ and $\mathbf{B}_{exp}$ are the PCA bases of identity and expression, each row of $\mathbf{R}$ is the offset of each vertex. %
Following Deng~\textit{et al.}~\cite{deng2019accurate}, we adopt the 2009 Basel Face Model~\cite{paysan20093d} for $\mathbf{\bar{S}}$ and $\mathbf{B}_{id}$, use expression bases $\mathbf{B}_{exp}$ of Guo~\textit{et al.}~\cite{guo2018cnn} built from Facewarehouse~\cite{cao2013facewarehouse}, resulting in $\alpha \in \mathbb{R}^{80}$, $\beta \in \mathbb{R}^{64}$. %
During implementation, we would add regularization to $\mathbf{R}$, which guarantees that the vertex level deformation does not deviate too far from the 3DMM deformation space. %



With the deformation space, we initialize 3D face $\mathbf{S}$ through facial landmark fitting. %
In this stage, the 3DMM parameters $\alpha, \beta$ are optimized to minimize the distance between the 3D landmarks of $\mathbf{S}$ and the detected 3D landmarks, while the vertex-level offset $\mathbf{R}$ is frozen to zero. We also penalize the squared sum of the optimized parameters, which imposes $\alpha, \beta$ to be close to zero. %

\subsection{3DMM Parameter Fitting}

Based on the 3D face $\mathbf{S}$ initialized from landmark fitting, %
in this stage, we further deform 3D face $\mathbf{S}$ to fit RGB-D videos captured from three synchronized RGB-D cameras. %
We rasterize $\mathbf{S}$ to obtain RGB images and depth images with a differentiable render~\cite{Laine2020diffrast}, and optimize the distance between rasterized images and ground truth images. %

More specifically, given the 3D face $\mathbf{S}$, %
we first transform $\mathbf{S}$ from the world space to the camera space of each camera. %
The 3D face under each camera space is denoted as $\mathbf{S}_{i}$ (where $i$ is the ID of each RGB-D camera). %
Based on $\mathbf{S}_{i}$, we then colorize $\mathbf{S}_{i}$ with 3DMM texture parameter $\delta \in \mathbb{R}^{80}$ and the spherical harmonic lighting parameter $\gamma \in \mathbb{R}^{27}$ following Deng~\textit{et al}~\cite{deng2019accurate}, yielding vertex color $\mathbf{C}_{i} \in \mathbb{R}^{n \times 3}$. %
Afterwards, the $\mathbf{C}_{i}$ and the z-axis of $\mathbf{S}_{i}$ are rasterized with differentiable render~\cite{Laine2020diffrast}, bringing out the rasterized RGB image $\mathbf{\hat{I}}_{\mathrm{rgb}}^{i}$ and depth image $\mathbf{\hat{I}}_{\mathrm{d}}^{i}$. %

For the 3D face reconstruction of single RGB-D video, $\mathbf{\hat{I}}_{\mathrm{rgb}}^{i}$ and $\mathbf{\hat{I}}_{\mathrm{d}}^{i}$ are optimized to be close to $\mathbf{I}_{\mathrm{rgb}}^{i}$ and $\mathbf{I}_{\mathrm{d}}^{i}$, where $\mathbf{I}_{\mathrm{rgb}}^{i}$ and $\mathbf{I}_{\mathrm{d}}^{i}$ are corresponding ground truth color and depth images  from the $i$-th camera. %
The optimization is conducted with the following loss function:
\begin{equation}
    \mathcal{L}(\alpha,\beta,\delta,\gamma) =  \mathcal{L}_{\mathrm{rgb}}+ \lambda_{\mathrm{d}}\mathcal{L}_{\mathrm{d}}+\lambda_{\mathrm{lm}}\mathcal{L}_{\mathrm{lm}}+\lambda_{\mathrm{p}}\mathcal{L}_{\mathrm{p}},
\end{equation}
where $\mathcal{L}_{\mathrm{rgb}}$ measures the RGB image distance, %
 $\mathcal{L}_{\mathrm{d}}$ measures the depth image distance. %
$\mathcal{L}_{\mathrm{rgb}}$ supervises 3D face to have visually consistent geometry and texture, while $\mathcal{L}_{\mathrm{d}}$ supervises 3D face to have accurate geometry. %
 Landmark loss $\mathcal{L}_{\mathrm{lm}}$ and prior term $\mathcal{L}_{\mathrm{p}}$ are also optimized in this stage. %
 
 For the optimization of multiple synchronized RGB-D videos, %
 the loss function $\mathcal{L}(\alpha,\beta,\delta,\gamma)$ of different cameras are summed together and optimized simultaneously. %
 $\alpha, \beta, \delta$ are shared among different cameras, while the lighting parameter $\gamma$ is unshared because different cameras have different imaging settings and lighting conditions. %

\subsection{Vertex-Level Fitting}

Through 3DMM parameter fitting, we have obtained a relatively accurate 3D face. %
However, the shape parameter $\alpha$ and expression parameter $\beta$ have limited capacity on representing subtle facial movements. %
To keep more accurate facial details, we further optimize vertex level deformation $\mathbf{R}$ together with the aforementioned parameters $\alpha, \beta, \delta, \gamma$. %

The objective of the vertex-level fitting stage is similar to that of the 3DMM parameter fitting: we still aim to minimize the distance between  $\mathbf{\hat{I}}_{\mathrm{rgb}}^{i}, \mathbf{\hat{I}}_{\mathrm{d}}^{i}$ and $\mathbf{I}_{\mathrm{rgb}}^{i}, \mathbf{I}_{\mathrm{d}}^{i}$. %
One main challenge of the vertex-level fitting is that the optimized parameter $\mathbf{R}$ has significantly more degree of freedoms ($3n$ degree of freedoms) compared with 3DMM parameter fitting. %
Regularizations are needed to prevent ill-posed optimization. %
We respectively incorporate edge regularization, laplacian smooth regularization, and offset penalty regularization. %
The edge regularization $\mathcal{L}_\mathrm{e}$ minimizes the edge length difference before and after the vertex-level fitting, which enforces local rigidness of deformation. %
The laplacian smooth regularization $\mathcal{L}_\mathrm{lap}$ minimizes the differential coordinate of each vertex with the  laplacian operator, which leads to local smoothness of deformation. %
The offset penalty regularization $\mathcal{L}_\mathrm{op}$ minimizes the L2-norm of each vertex offset, which guarantees that the offset do not deviate too far from the 3DMM deformation space. %
All of these regularizations are summed as vertex prior term $\mathcal{L}_\mathrm{vp}$:
\begin{equation}
    \mathcal{L}_\mathrm{vp} = \lambda_\mathrm{e}\mathcal{L}_\mathrm{e} + \lambda_\mathrm{lap}\mathcal{L}_\mathrm{lap} + \lambda_\mathrm{op}\mathcal{L}_\mathrm{op}.
\end{equation}
Combining $\mathcal{L}_\mathrm{vp}$ with the loss function in the 3DMM parameter fitting stage, %
we obtain the following loss:
\begin{equation}
    \mathcal{L}(\mathbf{R},\alpha,\beta,\delta,\gamma)=\mathcal{L}_{\mathrm{rgb}}+ \lambda_{\mathrm{d}}\mathcal{L}_{\mathrm{d}}+\lambda_{\mathrm{lm}}\mathcal{L}_{\mathrm{lm}}+\lambda_{\mathrm{p}}\mathcal{L}_{\mathrm{p}}+\mathcal{L}_\mathrm{vp}.
\end{equation}
When optimizing multiple synchronized RGB-D videos, the vertex level deformation $\mathbf{R}$ is shared across cameras. %

\subsection{Sequence Reconstruction}

\begin{figure*}[h]
  \centering
  \includegraphics[width=\linewidth]{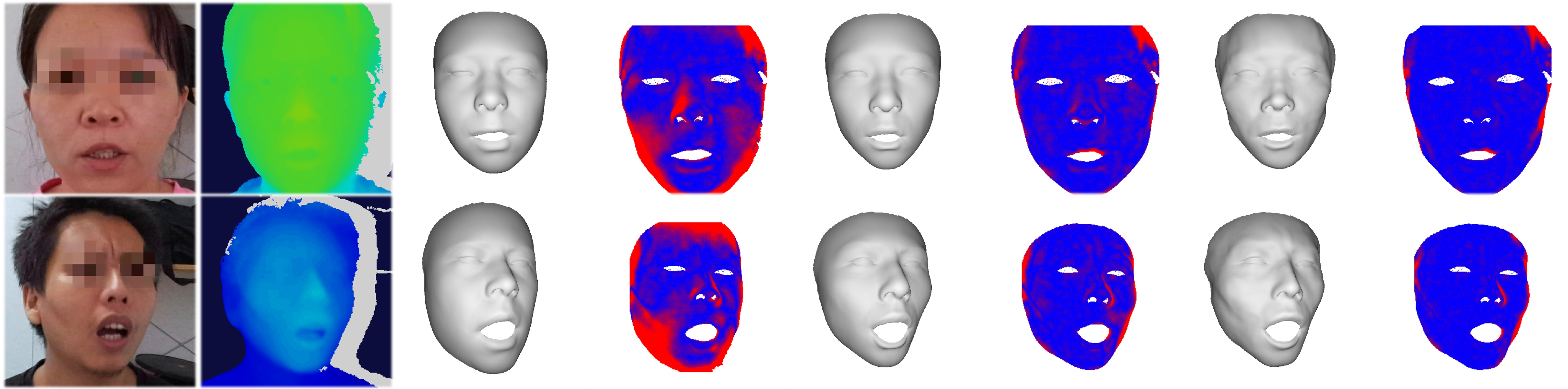}
  \caption{Visualizations of the reconstruction results. We respectively give the center-view of the RGB image, center-view of the depth image, the 3D mesh and the reconstruction error respectively after initialization, 3DMM fitting, and vertex-level fitting from left to right. Note that the reconstruction error calculates the distance between the depth image and the fitted mesh, the red color denotes higher reconstruction error and the blue color denotes lower error. The eyes of RGB facial images are mosaicked for privacy protection.  %
  }
  \label{fig:recon_vis}
\end{figure*}

For the 3D reconstruction of the whole video sequence, we do not conduct three-stage fitting for all frames. %
Instead, we only conduct a three-stage fitting for the first frame of the video. %
For the other frames, just the vertex-level fitting is applied with the parameters of the last frame as initialization. %
Such simplification enables faster reconstruction speed. %
For the optimization of the first video frame, we set the learning rate to $0.01$. The loss of landmark fitting is minimized for 100 iterations. %
The losses of 3DMM parameter fitting and vertex-level fitting are optimized for 500 iterations. %
For the optimization of the subsequent frames, we set the learning rate to $0.005$. %
The loss of vertex-level fitting is optimized for 200 iterations. %
For the weights of loss functions, we generally choose $\lambda_\mathrm{d}=2$, $\lambda_\mathrm{lm}=100$, $\lambda_\mathrm{p}=0.001$, $\lambda_\mathrm{e}=20$, $\lambda_\mathrm{lap}=20$, $\lambda_\mathrm{op}=0.01$. %
The adam optimizer~\cite{kingma2014adam} is adopted to optimize the aforementioned loss functions. %

\section{Dataset Observations}

\begin{figure*}[h]
  \centering
  \includegraphics[width=\linewidth]{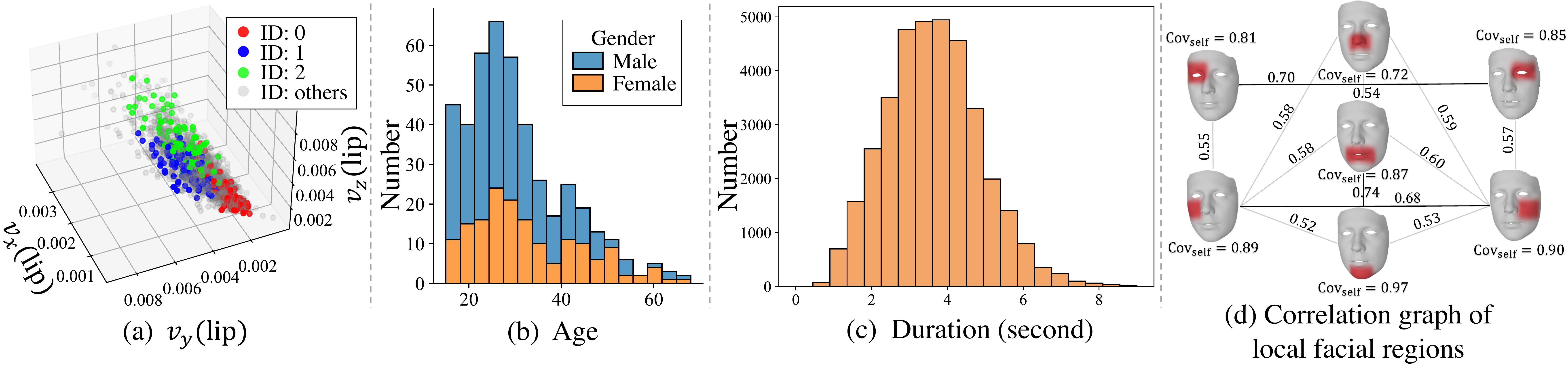}
  \caption{Statistics of the MMFace4D dataset. %
  (a) Scatter plot of $\boldsymbol{v}(\mathrm{lip})$ for each sequence of the MMFace4D, where $\boldsymbol{v}(\mathrm{lip})$ is the average vertex velocity of the lip region. %
  The sequences belonging to ID 0, 1, and 2 are colorized with red, blue, and green, while the other sequences are plotted with grey color. %
  (b) The histogram of subject age and subject gender. %
  (c) The histogram of sequence duration. %
  (d) The correlation graph of local facial regions. The local facial regions are colorized red. The $\mathrm{Cov_{self}}$ denotes the self-correlation inside the local region, and the weight of the edge denotes the correlation between the two regions.
  }
  \label{fig:statistics}
\end{figure*}

\subsection{Reconstruction Visualization}

To demonstrate the effectiveness of our reconstruction pipelines, %
we give qualitative visualizations of the reconstruction results in Figure~\ref{fig:recon_vis}. %
We observe that the pipeline of multi-stage fitting improves the reconstruction results progressively. %
Especially, after the vertex level fitting, subtle motions of the mouth, cheek, and eyebrow are reconstructed precisely. %
 More visualizations of the side-view image, and the rendering video of 3D mesh sequence are demonstrated in the supplementary. %


\subsection{Dataset Statistics}

Before analyzing the characteristics of MMFace4D, we first give an important statistic $\boldsymbol{v}(\cdot)$, which calculates the average vertex velocity along one particular axis for user-specified vertex sets. %
 For example, $\boldsymbol{v}_{x}(\mathrm{lip})$ calculates the average vertex velocity for vertices of the lip region along the x-axis. %
Based on $\boldsymbol{v}(\cdot)$, we conduct statistic analysis on the MMFace4D dataset to verify that our dataset has various talking styles, diversified actors, widely-distributed sequence durations, and expressive facial movements. %
Additionally, we verify that the dataset effectively captures both composite and regional characteristics. %


The talking style is well reflected in $\boldsymbol{v}(\mathrm{lip})$. %
As shown in Figure~\ref{fig:statistics}(a), we draw a scatter plot for some sequences of MMFace4D. %
The coordinate of each point in Figure~\ref{fig:statistics}(a) describes the $\boldsymbol{v}_{x}(\mathrm{lip})$, $\boldsymbol{v}_{y}(\mathrm{lip})$, and $\boldsymbol{v}_{z}(\mathrm{lip})$ value of each sequence. %
A few points of the scatter plot are colorized according to the subject ID of the sequence. %
From colorization, we observe that points of the same ID cluster together, while a gap exists among different identities. %
Such observation suggests that speech-independent factor such as taking style significantly impacts the distribution of facial movements, verifying the composite nature of facial animations. %

To investigate the impact of the regional nature on facial movements, we analyze motion correlation across different regions of the face. Motion correlation is used to quantify the degree of dependency between the movements of different facial regions. As shown in Figure~\ref{fig:statistics}(d), we first partition facial movements into small regions based on the facial mesh vertices to obtain the motion correlation, the local facial regions are colorized red. We then calculate the pairwise correlation coefficient along the spatial dimension between each pair of regions, resulting in a correlation matrix. Finally, we visualize this matrix as a connected graph. %
The $\mathrm{Cov_{self}}$ denotes the self-correlation inside the local region, and the weight of the edge denotes the correlation between the two regions. %
Edges with less than 0.5 correlation coefficient are removed. %
The correlation graph reveals that the motion correlation of facial regions is not uniformly distributed, with some regions being more correlated than others. 
For example, the mouth and chin regions, which are responsible for similar expressions, display a higher correlation, while the upper and lower parts of the face have a lower correlation.

 

In Figure~\ref{fig:statistics}(b) and Figure~\ref{fig:statistics}(c), %
we visualize the distributions of actor ages, genders, and sequence durations. %
In the MMFace4D dataset, the actor age ranges from 15 to 68,  %
the median age is 28, %
the male-female ratio is 1.65. %
The sequence duration ranges from 0.7 seconds to 11.4 seconds, %
the median duration is 3.6 seconds. %
Overall, the MMFace4D dataset contains diversified actors and widely-distributed sequence durations. %

MMFace4D dataset and previous datasets have significant differences in terms of average vertex velocity. %
For comparison, we average the vertex velocity of the lip region along all sequences and all axes. %
The $\boldsymbol{v}(\mathrm{lip})$ of the MMFace4D dataset is 0.0025, while the $\boldsymbol{v}(\mathrm{lip})$ of MeshTalk dataset~\cite{richard2021meshtalk} and VOCASET~\cite{cudeiro2019capture} are respectively 0.0016 and 0.0011. %
The MMFace4D dataset has faster vertex velocity compared with the VOCASET and MeshTalk dataset, %
which demonstrates that MMFace4D has more expressive and salient facial motions compared with previous datasets. %

The aforementioned characteristics of the MMFace4D dataset bring challenges for audio-driven 3D face animation. %
The 3D face animation model should on one hand be capable of synthesizing stylized and expressive talking faces that consider both composite and regional natures, on the other hand, generalize to different identities. %






\section{Audio2Face Framework}

\begin{figure*}[h]
  \centering
  \includegraphics[width=0.9\linewidth]{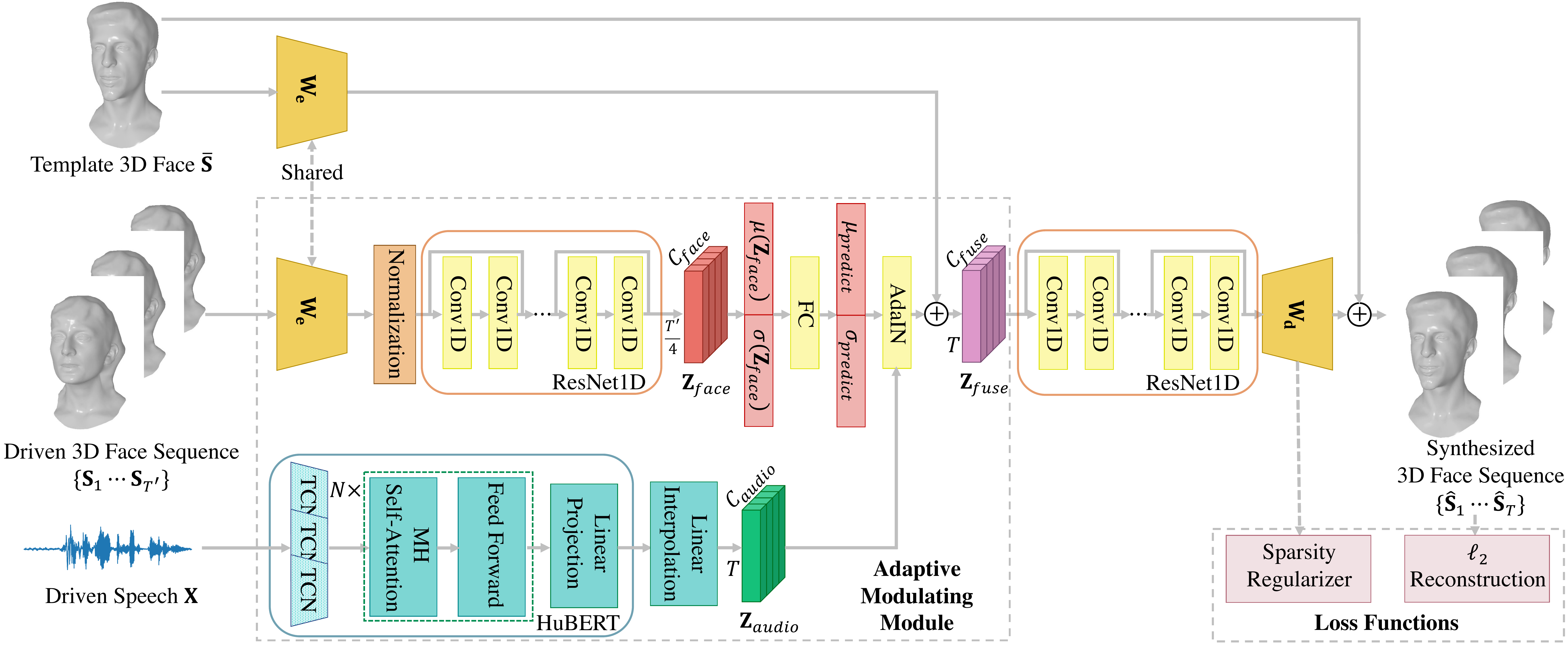}
  \caption{The overall framework of our method. The adaptive modulating module incorporates the composite nature of facial movements into the framework, while the sparsity regularizer interprets the regional nature of facial movements. %
  The overall backbone is non-autoregressive, which enables efficient training and inference.}
  \label{fig:framework_a2f}
\end{figure*}

In this section, we detailedly elaborate on how we integrate the regional and composite natures into the speech-driven 3D face animation framework. %
Afterwards, we introduce our non-autoregressive backbone, which preserves high-frequency details of facial animations and enables efficient inference. %

\subsection{Problem Formulation}

Before introducing the overall framework, we first formulate the problem of audio-driven 3D face animation in the presence of speech-independent factors. %
This task takes three inputs: a template 3D face $\mathbf{\bar{S}}$ of the target person, the driven speech $\mathbf{X}$ with duration $T$, and the driven 3D face sequence $\{\mathbf{S}_{1} \cdots \mathbf{S}_{T^{'}}\}$. %
The objective is to synthesize a sequence of 3D face animations $\{\mathbf{\hat{S}}_{1} \cdots \mathbf{\hat{S}}_{T}\}$. The synthesized animations have the same identity as $\mathbf{\bar{S}}$, are synchronized to the driven speech $\mathbf{X}$, and incorporate the speech-independent facial movements of $\{\mathbf{S}_{1} \cdots \mathbf{S}_{T^{'}}\}$. %

\subsection{Adaptive Modulating Module}

To incorporate the composite nature in 3D facial animation synthesis, combining speech-independent and speech-driven facial movements is imperative. %
Our proposed adaptive modulating module plays a crucial role in achieving this blend, as illustrated in Figure~\ref{fig:framework_a2f}. %
The module extracts the speech-independent representations from $\{\mathbf{S}_{1} \cdots \mathbf{S}_{T^{'}}\}$. %
More specifically, for each 3D face $\mathbf{S}_{i}$, we first normalize $\mathbf{S}_{i}$ by subtracting the mean face of $\{\mathbf{S}_{1} \cdots \mathbf{S}_{T^{'}}\}$. %
Formally:
\begin{equation}
    \mathrm{Norm}(\mathbf{S}_{i}) = \mathbf{S}_{i} - \frac{\sum_{i=1}^{t}\mathbf{S}_{i}}{t}.
\end{equation}
Normalization aims to extract solely the facial movement information while removing the identity information. %
The normalized $\mathrm{Norm}(\mathbf{S}{i})$ is then reduced to a low-dimensional feature vector using the encoding matrix $\mathbf{W}{e}$. The embedded vectors of 3D faces are concatenated into a sequence and input into a ResNet1D~\cite{he2016deep} encoder, producing latent face representations $\mathbf{Z}_{face}$ with a shape of $\frac{T^{'}}{4} \times C_{face}$, where $C_{face}$ is the channel number, and $\frac{T^{'}}{4}$ accounts for the downsampled embedded face vectors along the temporal dimension. %
Afterward, we extract the speech-independent facial movements from $\mathbf{Z}_{face}$. %
Different from the previous methods~\cite{richard2021meshtalk,peng2023emotalk,ji2021audio} that leverages cross reconstruction loss to extract speech-independent factors, 
we extract the speech-independent information by simply calculating the mean $\mu(\cdot)$ and standard deviation $\sigma(\cdot)$ of $\mathbf{Z}_{face}$ along the temporal dimension. %
Remarkably, this simple approach provides an effective representation of speech-independent information, as it captures the overall distribution of face animations while excluding the temporal information of $\mathbf{Z}_{face}$. %

Having obtained $\mu(\mathbf{Z}_{face})$ and $\sigma(\mathbf{Z}_{face})$, %
we now blend them with input speech signals. %
We first feed the input speech to the pretrained audio model~\cite{hsu2021hubert}, yielding the latent audio feature $\mathbf{Z}_{audio}$ with a shape of $T \times C_{audio}$, where $C_{audio}$ is the channel number of latent audio features. %
Then, we concat the template face embedding with $\mathbf{Z}_{audio}$, yielding $\mathbf{Z}_{concat}$. %
Afterwards, $\mu(\mathbf{Z}_{face})$ and $\sigma(\mathbf{Z}_{face})$ are used to modulate the mean and standard deviation of $\mathbf{Z}_{concat}$ on a global level. %
 Specifically, we map $\mu(\mathbf{Z}_{face})$ and $\sigma(\mathbf{Z}_{face})$ to $\mu_{predict}$ and $\sigma_{predict}$ with a linear layer, where $\mu_{predict}$ and $\sigma_{predict}$ have the same channel number as $\mathbf{Z}_{concat}$. Finally, we adjust $\mathbf{Z}_{concat}$ with a similar manner as AdaIN~\cite{huang2017arbitrary}:
\begin{equation}
    \mathbf{Z}_{fuse} = \sigma_{predict}(\frac{\mathbf{Z}_{concat} - \mu(\mathbf{Z}_{concat})}{\sigma(\mathbf{Z}_{concat})}) + \mu_{predict}.
\end{equation}
The acquired $\mathbf{Z}_{fuse}$ contains both speech driven and speech independent information. %

\subsection{Backbone}


Designing an efficient and effective backbone is also crucial for the task of audio-driven 3D face animation. In this section, we  illustrate how the backbone obtains the latent audio feature $\mathbf{Z}_{audio}$ and how the backbone generates $\{\mathbf{\hat{S}}_{1} \cdots \mathbf{\hat{S}}_{T}\}$ from $\mathbf{Z}_{fuse}$ and $\mathbf{\bar{S}}$. %

We utilize the pre-trained HuBERT model~\cite{hsu2021hubert} for audio encoding. Notice that we have compared HuBERT, wav2vec 2.0~\cite{baevski2020wav2vec}, Mel spectrogram, and DeepSpeech~\cite{amodei2016deep} features. Among these features, we observe that the HuBERT feature performs best. In our implementation, we extract the final layer of the HuBERT model and resample the output with the desired frame rate to obtain the latent audio feature $\mathbf{Z}_{audio}$. %

For the sub-modules within our backbone, we opt for the ResNet1D~\cite{he2016deep} structure over the Transformer structure. ResNet1D performs 1D convolution on the input feature vector sequence along the temporal dimension, exhibiting three key characteristics: (1) It imposes a strong inductive bias on the model architecture, aggregating information from temporally adjacent frames and reducing the required training data. (2) ResNet1D possesses a robust non-linear translation capability due to its stacked convolution layers. (3) Being a non-autoregressive and fully convolutional architecture, ResNet1D demands less computation cost and adapts to inputs of arbitrary size. %
These characteristics align well with the requirements of speech-driven 3D face animation. The strict temporal correspondence between input speech and 3D facial animation diminishes the need for complex time dependencies. When mapping speech to 3D facial animation, fusing temporal information from adjacent frames, as ResNet1D does, proves sufficient. Additionally, given the high heterogeneity between input speech and 3D facial animation, the potent non-linear translation capacity of ResNet1D is crucial for our task.


Based on the intuition above, we generate $\{\mathbf{\hat{S}}_{1} \cdots \mathbf{\hat{S}}_{T}\}$ from $\mathbf{Z}_{fuse}$ and $\mathbf{\bar{S}}$ with the ResNet1D decoder and decoding matrix $\mathbf{W}_{d}$. %
Notice that we have removed all of the downsampling layers in ResNet1D during the decoding process; all layers have a convolutional kernel with size 3 and stride 1. %
Such modification avoids synthesizing over-smooth animations and retains high-frequency details. %
Overall, both the encoder and the decoder of our backbone are non-autoregressive, %
thus can be trained in parallel and run efficiently during inference. The non-autoregressive design also allows for flexible and variable-length input sequences, making our model applicable to various applications. 

\subsection{Sparsity Regularizer}
\label{sec:sparse}

When decoding latent representations into 3D vertices using the decoding matrix $\mathbf{W}_{d}$, it is crucial to account for the regional nature of facial movements. Failure to consider this aspect may result in feature vectors that fail to capture subtle details in local regions. Our proposed approach addresses this issue without requiring manual intervention and is applicable to various 3D face templates.
To achieve this, we impose a constraint on $\mathbf{W}_{d}$ to ensure mutual exclusivity in its columns. Specifically, we require each column to exhibit sparsity, where if one row assigns a significant weight to a particular vertex, other rows should assign smaller weights to the same vertex. To enforce this constraint, we introduce a novel sparsity regularizer.
For clarity, let $\mathbf{W}_{d} \in \mathbb{R}^{m \times n}$, with each row denoted as $\mathbf{w}_{d}^{i}$, where $m$ is the feature number, and $n$ is the vertex number. We normalize each $\mathbf{w}_{d}^{i}$ and calculate the absolute values, denoted as $\lvert \mathbf{w}_{d}^{i} \rvert$. We require $\lvert \mathbf{w}_{d}^{i} \rvert \cdot \lvert \mathbf{w}_{d}^{j} \rvert$ to be small for $i \neq j$. Formally, the sparsity regularizer is defined as:
\begin{equation}
    \mathcal{L}_{reg} = \sum_{i=0}^{n} \sum_{j\neq i} \lvert \mathbf{w}_{d}^{i} \rvert \cdot \lvert \mathbf{w}_{d}^{j} \rvert.
\end{equation}
This sparsity constraint allows each element of the feature vector to focus on local facial regions, improving interpretability of learned weights and enhancing generalization capability.

\textbf{Training objectives.} For the overall optimization, we simultaneously optimize the \(\ell_2\) loss and the sparsity regularization loss. %
Formally:
\begin{equation}
\begin{split}
    \mathcal{L} & = \mathcal{L}_{\mathrm{\ell_{2}}} + \beta\mathcal{L}_{\mathrm{reg}} = \sum_{i = 1}^{T}||\mathbf{\hat{S}}_{i} - \mathbf{S}_{i}||_{2} + \beta\mathcal{L}_{reg}.
\end{split}
\end{equation}

\section{Experiments}

\begin{figure}[h]
  \centering
  \includegraphics[width=\linewidth]{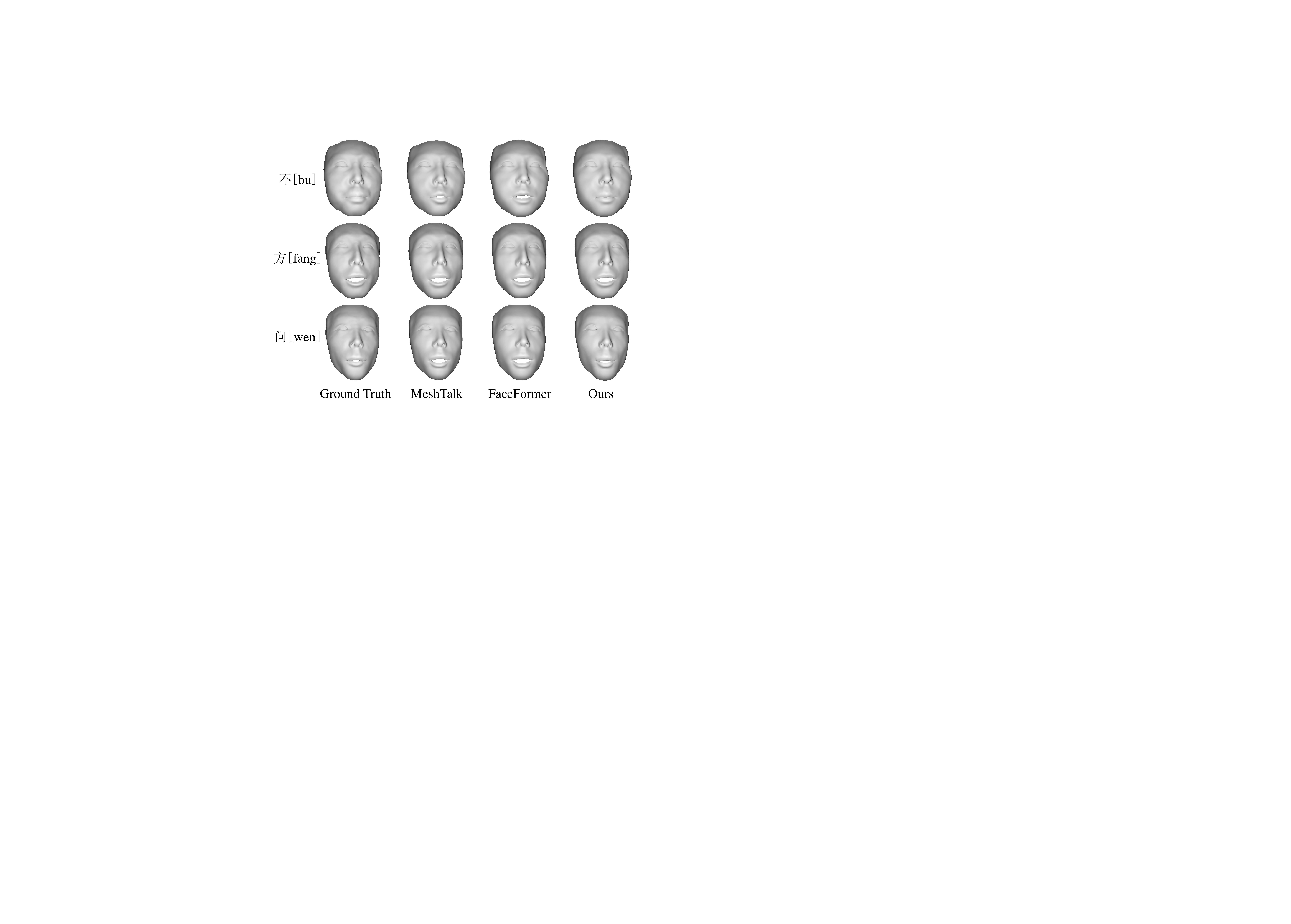}
  \caption{Qualitative comparison on the MMFace4D dataset. Each
row shows the facial animations when speaking different words (Pinyin labels provided in brackets for reference). }
  \label{fig:qualitative}
\end{figure}

\subsection{Experimental Settings}

\textbf{Implementation Details.} For the HuBERT model, we adopt the HuBERT-large configuration with 24 Transformer layers. %
For the ResNet1D model, we adopt the ResNet18 configuration. %
During training, we leverage the Adam optimizer with learning rate of $10^{-4}$. %
We train for $10^{6}$ iterations with a mini-batch size of 8 samples. %
In the implementation, the scaling coefficient $\beta$ of the regularization loss is set to $10^{-6}$. %

\textbf{Dataset Split.} We split the MMFace4D dataset into a training set and a test set by the identity of each actor, where the training set contains 400 actors, and the test set contains the remaining 31 actors. %

\textbf{Baseline Methods.} We conducted a comparative evaluation of our proposed framework with two state-of-the-art methods: MeshTalk~\cite{richard2021meshtalk} and FaceFormer~\cite{fan2022faceformer}. 

\textbf{Evaluation Metrics.} 
To evaluate lip synchronization, we employ two key metrics: maximal lip vertex error ($L_{\max}^{lip}$) and average lip vertex error ($L_{\mathrm{mean}}^{lip}$). $L_{\max}^{lip}$ measures the maximum Euclidean distance between lip region vertices in the synthesized and ground truth 3D faces, followed by averaging the error across frames. In contrast, $L_{\mathrm{mean}}^{lip}$ calculates the average distance between lip region vertices. %
For assessing synchronization in speech-independent movements, we utilize two additional metrics: maximal upper face error ($L_{\max}^{upper}$) and maximal face error ($L_{\mathrm{max}}^{face}$). These error metrics employ similar computation methods as $L_{\max}^{lip}$ but are applied to different facial regions. %

\begin{table}[]
    \centering
    \caption{Comparison with state-of-the-art methods on the MMFace4D dataset, lower denotes better for all metrics. Our method outperforms baseline methods in terms of both speech-driven movements and speech-independent movements. }
    \begin{tabular}{c||cccc}
    \midrule
                       & $L_{\max}^{lip}$       & $L_{\mathrm{mean}}^{lip}$       & $L_{\max}^{upper}$       & $L_{\mathrm{max}}^{face}$       \\ \midrule
    MeshTalk~\cite{richard2021meshtalk}           & 0.00208 & 0.00087 & 0.00360 & 0.00387 \\
    FaceFormer~\cite{fan2022faceformer}         & 0.00186 & 0.00073 & 0.00300 & 0.00330 \\ \midrule
    Ours w/o Composite & 0.00291 & 0.00121 & 0.00428 & 0.00480 \\
    Ours w/o Regional  & 0.00155 & 0.00056 & 0.00238 & 0.00270 \\
    \textbf{Ours}               & \textbf{0.00152} & \textbf{0.00054} & \textbf{0.00236} & \textbf{0.00268} \\ \midrule
    \end{tabular}
    \label{tab:quantative}
\end{table}

\subsection{Evaluations}

In Figure~\ref{fig:qualitative}, we present a qualitative comparison between our proposed method and baseline approaches. Each row illustrates the synthesized 3D faces corresponding to different input speeches. To ensure fair comparisons, all methods and input speeches adhere to the same talking style. %
Our method surpasses the baseline methods, showcasing more realistic and vivid 3D facial movements, particularly excelling in lip synchronization. Notably, our approach demonstrates more pronounced mouth opening and closing movements during the pronunciation of sounds like /b/ and /p/, as well as distinct pouting movements for sounds like /w/ and /v/. These nuanced expressions contribute to a heightened level of realism in the generated animations. %
In contrast, the baseline methods exhibit a jaw-flapping effect, leading to unnatural facial movements. %
The qualitative analysis emphasizes the superior performance of our proposed approach in generating lifelike facial movements under consistent talking styles. %

Table~\ref{tab:quantative} shows the quantitative results. %
Notably, the MeshTalk method shows suboptimal performance due to their limited capacity, particularly in accurately capturing nuanced facial expressions and speech-related variations. %
In contrast, the FaceFormer method exhibits superior generalization capacity, primarily attributed to the integration of pretrained audio models~\cite{baevski2020wav2vec,hsu2021hubert}. %
This integration allows FaceFormer to better adapt to different speech patterns and variations, enhancing its overall performance. %
Moreover, our approach significantly outperforms FaceFormer in both lip synchronization and speech-independent synchronization, benefiting from its comprehensive consideration of both composite and regional characteristics. %

In addition, our method is also computationally efficient due to the design of non-autoregressive architecture. %
Our method takes 0.01 seconds to synthesize 1-second 3D face sequences during inference, while the FaceFormer method takes 0.07 seconds. %
The efficiency of our method makes it practical for real-time applications such as video conferencing, telepresence, and gaming. %


\subsection{Ablation Study}

\begin{figure}[h]
  \centering
  \includegraphics[width=0.9\linewidth]{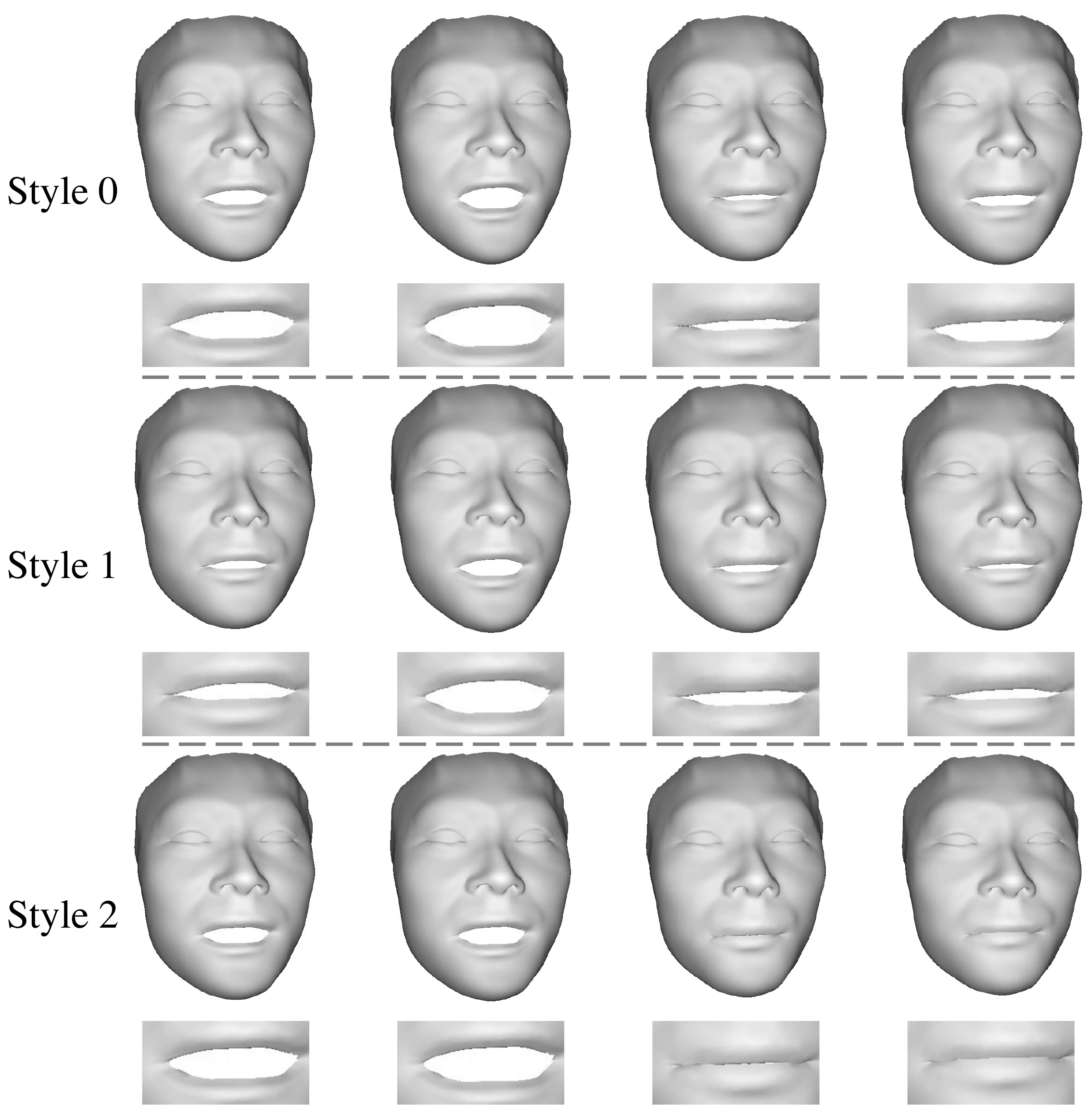}
  \caption{Stylized 3D face animations driven by one input speech. We amplify the mouth region of each face. The style is mainly reflected in mouth close and open.}
  \label{fig:style}
\end{figure}

We conducted ablation studies to evaluate the effectiveness of considering composite and regional natures in synthesizing 3D face animations. Specifically, we performed experiments by selectively removing the composite or regional nature from our model architecture and synthesizing speech-driven 3D face animations using the modified models. The quantitative results of these experiments are reported in Table~\ref{tab:quantative}. %

The results underscore the detrimental impact of removing the regional nature from our model on its overall performance. The regional nature played a crucial role in enabling our model to concentrate on specific details of facial movements, particularly around the lips. Although the enhancement of these details exerted a mild negative influence on the evaluation metrics, it contributed to the fidelity of our synthesized animations. %
Meanwhile, the removal of the composite nature from our model resulted in a substantial decline in performance, particularly in the synthesis of speech-independent facial movements. The composite nature was instrumental in empowering our model to capture global patterns and overarching trends in facial movements that remain independent of speech cues. Our findings unequivocally advocate for the importance of both regional and composite natures in the synthesis of speech-driven 3D face animations.

\begin{figure}[h]
  \centering
  \includegraphics[width=\linewidth, height=3cm]{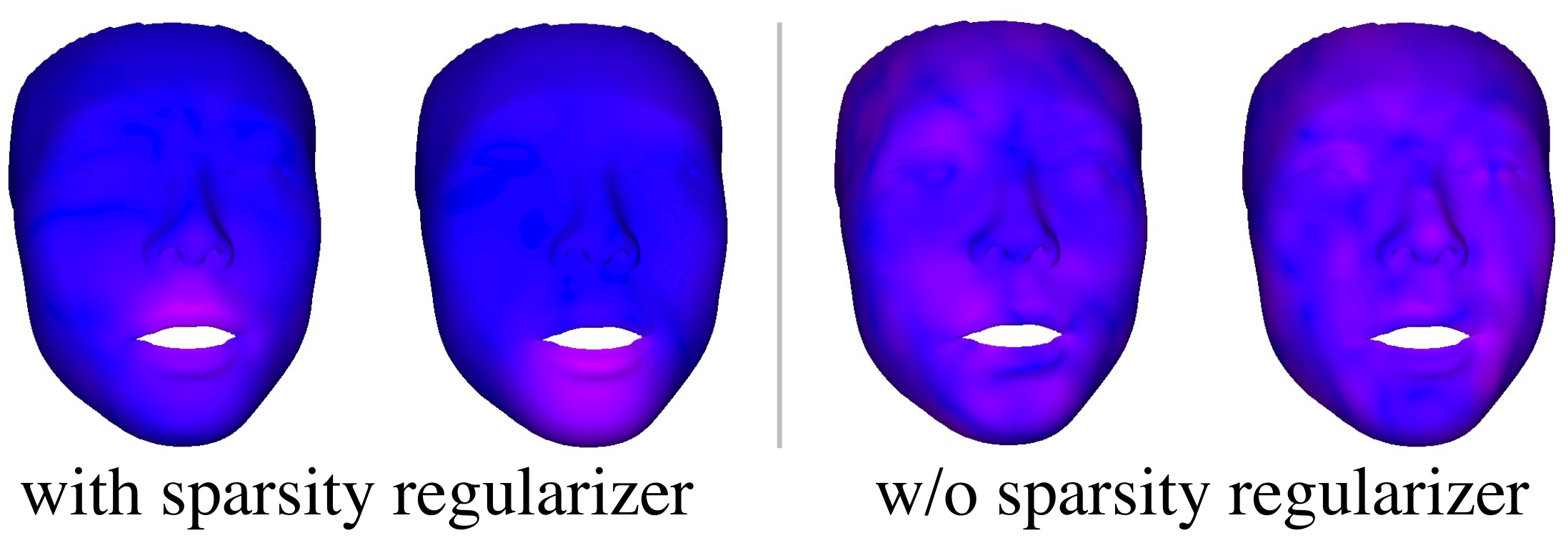}
  \caption{The activated region for each element of the motion features. Blue regions indicate no influence, while red regions denote high influence. }
  \label{fig:region}
\end{figure}

We have also created visualizations to showcase the effectiveness of our method in synthesizing composite and regional facial movements. %
Figure~\ref{fig:style} shows that our method successfully generates diverse talking styles for single-driven speech input, highlighting the ability to effectively capture the nuances and variations in facial expressions characteristic of different speaking styles. %
To further demonstrate the effectiveness of our method, we also draw Figure~\ref{fig:region} to illustrate the impact of our proposed sparsity regularizer, which enforces each facial feature element to focus on the local region of mesh vertices. By incorporating this sparsity regularizer, our method can identify and extract interpretable regions for synthesizing facial movements, leading to more natural and accurate results. Meanwhile, when the sparsity regularizer is removed, the activated regions of motion features spread across the face. %

Overall, our ablation study provides empirical evidence for the effectiveness of considering both composite and regional natures in synthesizing 3D face animations. By combining the two natures, our model achieves superior performance.

\section{Conclusion}

In this paper, we present a large-scale multi-modal 4D~(3D sequence) face dataset named MMFace4D. %
MMFace4D features the following two main properties: %
(1) diversified corpus and actors, %
(2) synchronized speech audio and high-fidelity 3D animation with our multi-stage reconstruction pipeline. %
Through extensive data observations on MMFace4D, we showcase the dataset's richness in various talking styles, expressive facial motions, and a broad spectrum of actors. Leveraging MMFace4D as a foundation, we propose a highly effective audio-driven 3D face animation method. This method takes into account both the composite and regional natures of facial movements, enabling non-autoregressive generation. Notably, our approach outperforms state-of-the-art autoregressive methods, achieving superior performance with faster inference. %
The introduction of MMFace4D represents a substantial advancement in the field of audio-driven 3D face animation, transitioning from previous small-scale scenarios to large-scale scenarios. This transition facilitates the training of high-fidelity, expressive, and generalizable face animation models. We anticipate that the release of MMFace4D will catalyze further research in this domain. %



\textbf{Ethics Considerations}: Each actor of the MMFace4D dataset participates in the recording voluntarily. %
For the issue of privacy protection, all demonstrated RGB face images are mosaicked. %
The MMFace4D dataset aims to promote positive technologies. %
However, we also acknowledge that both the dataset and the technology of audio-driven 3D face animation have a risk of being misused. %
Thus, we hope to raise the awareness of the public and develop forgery detection technology to prevent such misuse. %


\bibliographystyle{IEEEtran}
\bibliography{egbib}

\end{document}